\documentclass{article} 
\usepackage{nips15submit_e,times}
\let\chapter\section
\usepackage{algorithm2e}
\usepackage{hyperref}
\usepackage{url}
\usepackage{amsmath}
\usepackage{amsthm}
\usepackage{amssymb}
\usepackage{graphicx}
\usepackage{xspace}
\usepackage{tabularx}
\usepackage{multicol}
\usepackage{multirow}
\usepackage{url}
\usepackage[numbers,comma,square]{natbib}
\usepackage{wrapfig}
\usepackage[utf8]{inputenc}
\usepackage[normalem]{ulem}
\usepackage{xcolor}

\newcommand{\bmx}[0]{\begin{bmatrix}}
\newcommand{\emx}[0]{\end{bmatrix}}

\newcommand{\vect}[1]{\mathbf{#1}}

\newcommand{\vh}[0]{\vect{h}}

\newcommand{\vx}[0]{\vect{x}}
\newcommand{\vz}[0]{\vect{z}}

\newcommand{\vs}[0]{\vect{s}}

\graphicspath{{./figures/}}

\title{Detecting Interrogative Utterances\\with Recurrent Neural Networks}

\author{
Junyoung Chung\thanks{Work done while the author was at Microsoft.}\\
Universit\'{e} de Montr\'{e}al\\
Montr\'{e}al, Canada\\
\texttt{junyoung.chung@umontreal.ca}\\
\And
Jacob Devlin\;\;\;\;Hany Hassan Awadalla\\
Microsoft Research\\
Redmond, USA\\
\texttt{\{jdevlin, hanyh\}@microsoft.com}\\
}

\nipsfinalcopy 

\begin{document}

\maketitle

\begin{abstract}
In this paper, we explore different neural network architectures that can predict
if a speaker of a given utterance is asking a question or making a statement.
We compare the outcomes of regularization methods that are popularly used to train deep neural networks
and study how different {\it context} functions can affect the classification performance.
We also compare the efficacy of gated activation functions that are favorably used in recurrent neural networks
and study how to combine multimodal inputs.
We evaluate our models on two multimodal datasets: MSR-Skype and CALLHOME.
\end{abstract}

\section{Introduction}
\label{sec:intro}
Spoken language understanding is a long-term goal of machine learning and potentially has a huge impact in practical applications.
However, the difficulty of processing speech signals itself is a bottleneck, for instance, the core part of speech translation has to be processed in the text domain.
In other words, a failure of capturing the key features in the speech signals can lead the next applications into unexpected results.

Identifying whether a given utterance is a question or not can be one of the key features in applications such as speech translation.
Unfortunately, a speech recognition system is likely to fail achieving two goals at a same time:
(1) extract text sequences from the input utterances, (2) detect questions.
We can think of a question detection system that works independently and unburdens the load of the
speech recognition system~\citep{yuan2005detection,metzler2005analysis,boakye2009any,wang2010exploiting,bazillon2011speaker}.
Later, an annotation of being a question can form a set with the output of the speech recognition system
and handed over to the machine translation system.

Previous studies have focused on using hand-designed features and classifiers such as support vector machines (SVMs)~\citep{metzler2005analysis,wang2010exploiting} or tree-based classifiers~\citep{yuan2005detection,boakye2009any,bazillon2011speaker}.
The classifiers used in these systems are shallow and simple, but there are considerable efforts on designing features based on domain knowledge. 
However, there is no guarantee that these hand-designed features are optimal to solve the problem.

In this study, we will let the model to learn the features from the training examples and the objective function.
We propose a recurrent neural network (RNN) based system with various model architectures that can detect questions using multimodal inputs.
Our question detection system runs as fast as other real-time systems at the test time,
receives multimodal inputs and returns a scalar score value $\hat{y}\in[0, 1]$.
We evaluate our models on two multimodal datasets, which consist with pairs of text transcripts and audio signals.
Our experiments reveal what types of {\it context} functions, regularization methods,
state transition functions of RNNs and data domains are helpful in RNN-based question detection systems.

\begin{table}[t]
    \label{tab:question_type}
    \caption{Types of questions}
    \vfill
    \centering
    \begin{tabular}{| c | c |}
        \hline
        & Examples\\
        \hline
        Yes-No & Did you attend the meeting?\\
        \hline
        {\it wh}-words & Where have you been?\\
        \hline
        Declarative & You are at the meeting?\\
        \hline
    \end{tabular}
\end{table}

\section{Background}
\subsection{Types of Questions}
\label{subsec:type_of_questions}
Questions can have different canonical forms, and they are usually not standardized.
However, we can divide the questions into three groups based upon some criteria.
Table~\ref{tab:question_type} shows an example from each group.
We note that declarative questions are rather unclear to differentiate from non-question statements by
looking into their canonical forms because they usually do not contain any {\it wh}-words.
However, audio signals might contain the features that can be useful when making predictions on this kind of examples,
where a question usually contains a rising pitch at the end of the utterance.

\subsection{Neural Networks}
An RNN can process a sequence $\vx=(\vx_1,\vx_2,\dots,\vx_T)$ by recursively applying a transition function
$g$ to each symbol:
\begin{align}
    \vs_t=g(\vx_t,\vs_{t-1}),
\end{align}
where $g$ is usually a deterministic non-linear transition function.
$g$ gains extra strength to capture long-term memories when implemented with gated activation functions~\citep{chung2014empirical}
such as long short-term memory~\citep[LSTM,][]{Hochreiter+Schmidhuber-1997} or
gated recurrent unit~\citep[GRU,][]{cho2014emnlp}.
We can add more hidden layers in advance or subsequent to the RNN to increase the capacity of the model such that:
\begin{align}
    \vz_t=h(g(f(\vx_t),\vs_{t-1})),
\end{align}
where, a sequence $\vz=(\vz_1, \vz_2,\dots,\vz_T)$ is the transformed feature representation of the input sequence $\vx$, and
$f$ and $h$ are additional hidden layers.
Instead of using the whole $\vz$, we can apply a context function $c$ to reduce the dimensionality
and take only the abstract information out of $\vz$. The context function $c$ can be either defined as introduced in \citep{cho2014emnlp}:
\begin{align}
    \label{eq:context_base}
    c_1(\vz)=\vz_T,
\end{align}
or as introduced in \citep{bahdanau2014neural}:
\begin{align}
    \label{eq:context_attention}
    c_2(\vz)=\sum_{t=1}^T\alpha_t\vs_t,
\end{align}
where $\alpha_t$ is the weight of each annotation $\vh_t$.
$c(\vz)$ can be used as the learned features for the logistic regression classifier:
\begin{align}
    \hat{y}=\sigma(c(\vz)),
\end{align}
where $\sigma$ is a notation of a sigmoid function.

In this study, we implement $f$ and $h$ with deep neural networks (DNNs) using fully-connected layers
and rectified linear units~\citep{nair2010rectified} as non-linearity,
$g$ with a recurrent layer using either GRU or LSTM as the state transition function,
and the context function $c$ with either Eq.~\eqref{eq:context_base} or Eq.~\eqref{eq:context_attention}.

\section{Proposed Models}
We take a neural network based approach where we can stack multiple feedforward and recurrent layers to
learn hierarchical features from the training examples and the objective function via stochastic
gradient descent.

\begin{figure*}[ht]
    \centering
    \includegraphics[width=0.6\columnwidth]{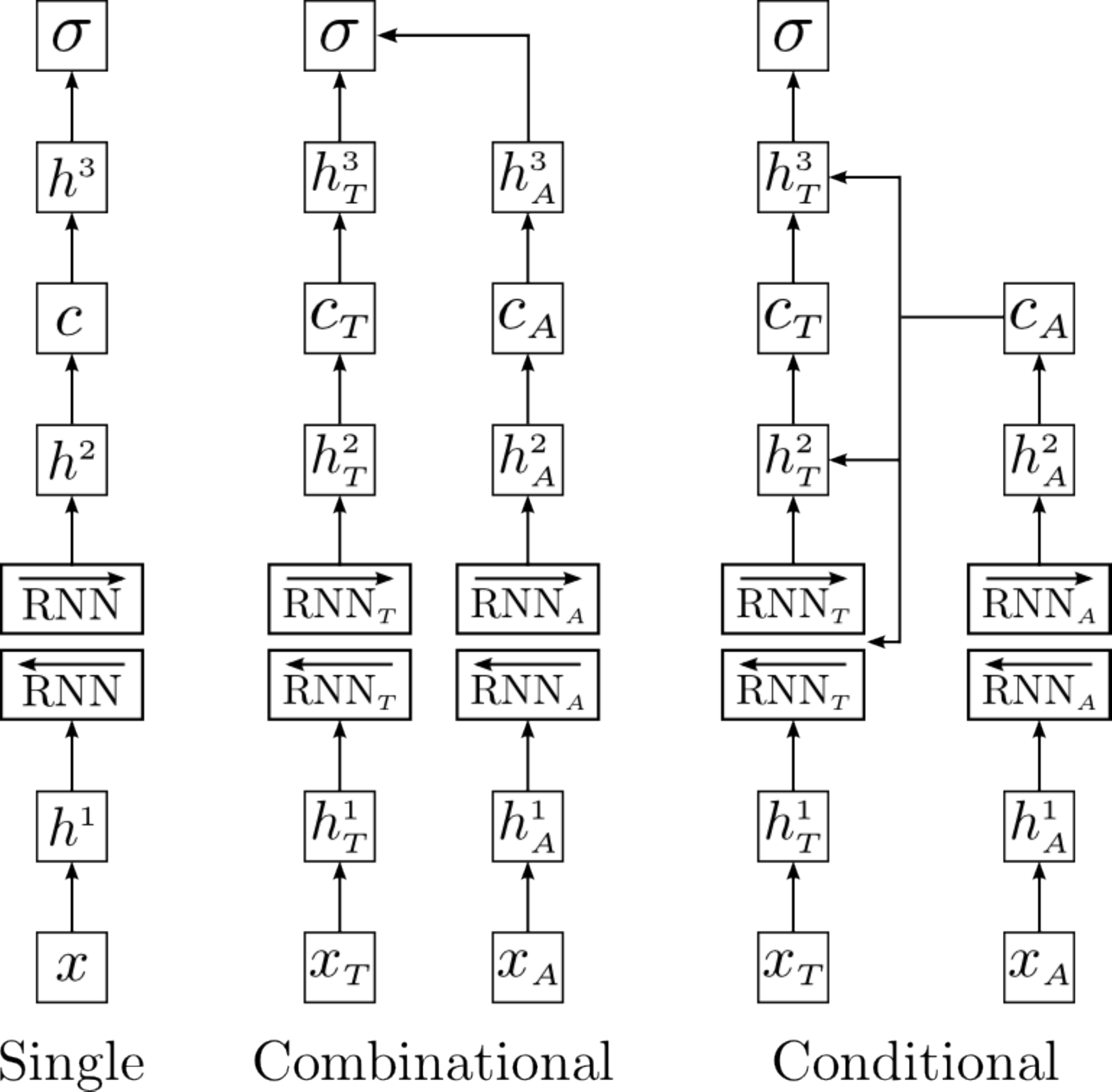}
     \caption{Graphical illustration of each group, note that the context function $c$ can be implemented
              as either Eq.~\eqref{eq:context_base} or Eq.~\eqref{eq:context_attention}.
              $x$ of `Single' model can be either a text input or an audio input.
              `Combinational' and `Conditional' models take both inputs, where the subscript $_T$ stands for text source
              and $_A$ stands for audio source.
              The topmost blocks with $\sigma$ indicate the logistic regression classifiers.
              Each layer has $200$ hidden units.
              The training objective is to minimize the average negative log-likelihood of the training examples.
              We implement the RNN with its bidirectional variant~\citep{schuster1997bidirectional}.}
    \label{fig:models}
\end{figure*}

We consider two types of inputs, which are text transcripts and audio signals of utterances.
Depending on what types of inputs are used,
we can divide the models into three groups: (1) receive only text inputs, (2) receive only audio inputs and (3) receive both inputs.
When a model receives both inputs as (3), we can think of a simple but naive way of combining two different features as shown as `Combinational' in Fig.~\ref{fig:models}.
For a model in each group, it can choose the context function to become as either Eq.~\eqref{eq:context_base} or Eq.~\eqref{eq:context_attention},
so the number of combinations becomes six.
However, there is another model, receives both inputs, uses Eq.~\eqref{eq:context_attention} as the context function,
but uses a different way of combining two features that is depicted as `Conditional' in Fig.~\ref{fig:models}.

For each model in each group, we train it with three different ways: (1) without any regularization methods,
(2) use dropout~\citep{srivastava2014dropout} and (3) use batch normalization (BN)~\citep{ioffe2015batch}
(note that we are not the first to apply batch normalization to a neural network architecture that contains an RNN~\citep{laurent2015batch}).
However, there is another diversity, the state transition function of the RNN hidden state, which can be implemented either as a GRU or an LSTM.
Therefore, for each model, there are six different candidates to compare with.
Recall that we have seven different models, each model has six different variations,
there is a total of $42$ candidates to be tested on two datasets that are MSR-Skype and CALLHOME.

\section{Experiment Settings}
\subsection{Datasets}
\paragraph{MSR-Skype}
MSR-Skype dataset contains $18,006$ examples given as text-audio pairs, and the proportion of positive and negative examples are well-balanced.
Each example is an utterance, which is segmented manually.
We only use examples that contain $3$ to $25$ words to train the models.
We use $80\%$ of the examples as a training set and reserve $20\%$ of the examples to validate and evaluate the models.

\paragraph{CALLHOME}
We use a subset of the original CALLHOME, where the text transcripts are created by human annotators.
There are $2,528$ examples given as text-audio pairs.
Utterances are segmented manually, and the train/validation/test splits are divided as same as the MSR-Skype dataset.

\paragraph{Preprocessing}
For the text data, we remove punctuations, commas, question marks, exclamation marks
to prevent the model from making decisions based on these special tokens.
We do not consider pretraining word representation vectors with external datasets, however, they are learned jointly with the objective function
during the training procedure.
Therefore, $h^{\scriptscriptstyle{1}}$ in `Single' (only when $x$ is text data)
and $h^{\scriptscriptstyle{1}}_{\scriptscriptstyle{T}}$ in `Combinational' and `Conditional' become
continuous vector representations of the words (in this case, we do not apply non-linearity).
We built the dictionary from MSR-Skype and CALLHOME, which contains 13,911 vocabularies.

We extract MFCC from the raw audio signals with $40ms$ frame duration, and $15ms$ overlap.
The lengths of the audio sequences (after extracting MFCC) could be significantly longer than the text sequences, therefore, in order to reduce the number of timesteps,
we concatenate four frames into one chunk and treat it as a single frame.

\subsection{Results}

\begin{table}[t]
    \caption{Test F1 score of the models trained on MSR-Skype.
             First two columns use neither dropout (shortened as D) nor BN.}
    \vspace{0.5cm}
    \vfill
    \centering
    \begin{tabular}{| c || c | c | c | c | c | c |}
        \hline
        & GRU & LSTM & GRU, D & LSTM, D & GRU, BN & LSTM, BN \\
        \cline{2-7}
        \hline
        \hline
        text, $c_1$        & 88.8 & 88.6 & 89.1 & 89.1 & 90.6 & 90.2 \\
        \hline
        text, $c_2$        & 89.5 & 88.9 & 88.9 & 88.7 & 90.8 & 90.5 \\
        \hline
        audio, $c_1$       & 77.3 & 73.7 & 79.2 & 77.0 & 75.3 & 71.2 \\
        \hline
        audio, $c_2$       & 77.2 & 77.6 & 80.7 & 81.2 & 76.5 & 76.8 \\
        \hline
        combination, $c_1$ & 89.2 & 88.9 & 89.1 & 88.8 & 90.5 & 90.3 \\
        \hline
        combination, $c_2$ & 89.7 & 88.9 & 89.2 & 89.0 & 90.9 & 91.0 \\
        \hline
        condition,   $c_2$ & 90.0 & 89.8 & 90.0 & 90.1 & 90.7 & 90.1 \\
        \hline
    \end{tabular}
    \label{tab:msr_skype_set_a}
\end{table}

Table~\ref{tab:msr_skype_set_a} shows the results of the models trained on MSR-Skype dataset.
We can observe a few tendencies in the obtained results depending on what kind of variations are applied to the models ($c_1$ or $c_2$,
GRU or LSTM, dropout or batch normalization and types of inputs).

In general, using both input sources are helpful, but the advantage is not that impressive when batch normalization is used for training.
The lengths of the audio sequences are usually longer than the text sequences, and attention mechanism ($c_2$)~\citep{bahdanau2014neural} is known to be a nice solution to deal with long sequences.
Therefore, when the model can only take audio inputs, $c_2$ is a better option than $c_1$.

Dropout will help in most cases, however, when using both input sources, the performance does not improve that much.
In fact, the performance gets worse than the models, which do not use dropout.
We assume that the optimization problem becomes difficult with dropout when the models receive both input sources, hence, in this case we need more care in using dropout.
Batch normalization improves the performance with a huge gap for the models that receive text source as inputs.
However, batch normalization does not help the models that can only receive audio inputs.
The best performance is achieved by a model that receives both input sources (combinational), uses $c_2$ as context function, uses batch normalization for training
and uses LSTM as the state transition function of the RNN.


Table~\ref{tab:callhome} shows the result of each model trained on CALLHOME.
We can observe that $c_2$ helps the models that only take audio inputs,
and batch normalization improves the performance of the models that includes text source as their inputs.
The best performance is achieved by a model that takes both input sources (combinational), uses $c_1$ as context function, uses batch normalization for training
and uses GRU as the state transition function of the RNN.

\begin{table}[t]
    \caption{Test F1 score of the models trained on CALLHOME.
             First two columns use neither dropout (shortened as D) nor BN.}
    \vspace{0.5cm}
    \vfill
    \centering
    \begin{tabular}{| c || c | c | c | c | c | c |}
        \hline
        & GRU & LSTM & GRU, D & LSTM, D & GRU, BN & LSTM, BN \\
        \cline{2-7}
        \hline
        \hline
        text, $c_1$        & 81.3 & 81.1 & 81.4 & 80.6 & 82.9 & 82.7 \\
        \hline
        text, $c_2$        & 81.0 & 80.7 & 81.2 & 81.5 & 83.5 & 83.0 \\
        \hline
        audio, $c_1$       & 70.9 & 70.0 & 72.2 & 72.4 & 67.0 & 66.3 \\
        \hline
        audio, $c_2$       & 70.8 & 71.8 & 72.5 & 73.4 & 69.2 & 67.8 \\
        \hline
        combination, $c_1$ & 83.1 & 82.7 & 82.6 & 82.8 & 84.6 & 84.0 \\
        \hline
        combination, $c_2$ & 83.0 & 82.5 & 82.7 & 82.4 & 84.1 & 83.6 \\
        \hline
        condition,   $c_2$ & 83.8 & 83.9 & 83.9 & 83.9 & 82.4 & 83.7 \\
        \hline
    \end{tabular}
    \label{tab:callhome}
\end{table}

\begin{table}[t]
    \caption{Test F1 score of the models trained on MSR-Skype and tested on variable-length sequences.}
    \vspace{0.5cm}
    \vfill
    \centering
    \begin{tabular}{| c || c | c | c | c |}
        \hline
        & text, $c_2$ & audio, $c_2$ & combination, $c_2$ & condition, $c_2$ \\
        \cline{2-5}
        \hline
        \hline
        Short Sequences & 77.6 & 64.6 & 78.3 & 79.3 \\
        \hline
        Intermediate Sequences   & 90.1 & 82.3 & 90.5 & 91.2 \\
        \hline
        Long Sequences  & 80.3 & 69.4 & 82.8 & 84.0 \\
        \hline
    \end{tabular}
    \label{tab:length}
\end{table}

In Table~\ref{tab:length}, we test our models on sequences with different lengths.
We use the same models that were trained on MSR-Skype, without any regularization methods.
The sequences are divided into three groups depending on the number of words contained in each sequence.
Short sequences have less than $5$ words, long sequences have more than $20$ words, and intermediate sequences contain $5$ to $20$ words.
We observe that the models achieve the best performance on intermediate sequences, and the models tend to do better jobs on short and long sequences when the inputs contain text source.
The performance degradations on short or long sequences compared to intermediate sequences are smaller when we use both input sources
(see `combination' and `condition', especially models lose less performance against long sequences).

\begin{table}[t]
    \caption{Examples of predicted scores on declarative questions.}
    \vfill
    \centering
    \begin{tabular}{| c || c | c | c | c|}
        \hline
        Test Examples& text, $c_2$ & audio, $c_2$ & combination, $c_2$ & condition, $c_2$ \\
        \hline
        \hline
        any other questions? & 0.44 & 0.98 & 0.72 & 0.84 \\
        \hline
        and your cats? & 0.63 & 0.93 & 0.97 & 0.72 \\
        \hline
        oh, the bird? & 0.42 & 0.83 & 0.77 & 0.72 \\
        \hline
    \end{tabular}
    \label{tab:result_anal}
\end{table}

Table~\ref{tab:result_anal} shows some test examples that neither contain {\it wh}-words nor have canonical form of questions,
which we have already introduced as declarative questions in Sec.~\ref{subsec:type_of_questions}.
In this kind of questions, there are usually rising pitches at the end of the audio signals.
For the models, which receive the audio source as inputs,
can benefit from having audio information as shown in Table~\ref{tab:result_anal} (see `audio', `combination' and `condition').
For the models, which receive only text source as inputs, do not have relevant information to guess whether the given utterances are questions or not.

The predicted scores from the models using both inputs are sometimes less than the scores from the model using only audio inputs.
We assume that these models have to make compromise between the text features and audio features when these two are in conflict.
However, given the training objective, it is difficult to expect that the models will completely ignore one of the features, instead,
the models will tend to learn more smooth decision boundaries.

\section{Conclusion}
We explore various types of RNN-based architectures for detecting questions in English utterances.
We discover some features that can help the models to achieve better scores in the question detection task.
Different types of inputs can complement each other, and the models can benefit from using both text and audio sources as inputs.
Attention mechanism ($c_2$) helps the models that receive long audio sequences as inputs.
Regularization methods can help the models to generalize better, however, when the models receive multimodal inputs,
we need to be more careful on using these regularization methods.

\section*{Acknowledgments}
\label{sec:ack}
The authors would like to thank the developers of Theano~\citep{Bastien-Theano-2012}.
We acknowledge the support of the following agencies for research funding and
computing support: Microsoft, NSERC, Calcul Qu\'{e}bec, Compute Canada,
the Canada Research Chairs and CIFAR.
\small

\newpage
\bibliography{strings,strings-shorter,aigaion,ml,junyoung}
\bibliographystyle{abbrvnat}

\end{document}